\documentclass[letterpaper, 10 pt, conference]{ieeeconf}

\IEEEoverridecommandlockouts %
\usepackage{times}

\usepackage{multicol}
\usepackage{hyperref}
\usepackage{booktabs}
\usepackage{graphicx}
\usepackage{float}
\usepackage{amsmath, amssymb}
\usepackage{amsfonts}
\usepackage{bm}
\usepackage{algpseudocode}
\usepackage{algorithm}
\usepackage{multirow}
\usepackage{overpic}
\usepackage{wrapfig}

\usepackage[normalem]{ulem} %

\usepackage[noadjust]{cite}

\newcommand{\xv}{\vec{x}} %
\newcommand{\gv}{\vec{g}} %
\newcommand{\rv}{\vec{r}} %
\newcommand{\yv}{\vec{y}} %
\newcommand{\vv}{\vec{v}} %
\newcommand{\ev}{\vec{e}} %
\newcommand{\mv}{\vec{m}} %

\renewcommand{\ij}{\overrightarrow{ij}}

\newcommand{\rdgnn}{RD-GNN}
\newcommand{\pce}{\phi_{PC}}
\newcommand{\actionenc}{\phi_{A}}
\newcommand{\graphenc}{\phi_{G}}
\newcommand{\ginput}{\xv^{I}}
\newcommand{\glatent}{\xv^{L}}
\newcommand{\glatentprime}{\xv^{L^\prime}}

\renewcommand{\vec}{\mathbf}

\usepackage{subcaption}
\usepackage[skip=0pt,font=small,labelfont=bf]{caption}
\usepackage{amsmath}
\DeclareMathOperator*{\argmax}{arg\,max}

\newcommand{\shrinka}{\def\baselinestretch{0.993}\large\normalsize}
\usepackage{color}
\definecolor{international_orange}{RGB}{240, 74, 0}

\title{Planning for Multi-Object Manipulation with \\Graph Neural Network Relational Classifiers}

\author{%
  Yixuan Huang$^{1}$,
  Adam Conkey$^{1}$,
  and Tucker Hermans$^{1, 2}$ 
  \thanks{$^{1}$%
    University of Utah.
    $^{2}$%
    NVIDIA.
    \protect\url{yixuan.huang@utah.edu},
    \protect\url{adam.conkey@utah.edu}, and
    \protect\url{thermans@cs.utah.edu}
  }%
}

\begin{document}
\input{overview_fig}

\maketitle

\begin{abstract}
  Objects rarely sit in isolation in human environments. As such, we'd like our robots to reason about how multiple objects relate to one another and how those relations may change as the robot interacts with the world.
  To this end, we propose a novel graph neural network framework for multi-object manipulation to predict how inter-object relations change given robot actions. Our model operates on partial-view point clouds and can reason about multiple objects dynamically interacting during the manipulation. By learning a dynamics model in a learned latent graph embedding space, our model enables multi-step planning to reach target goal relations.
  We show our model trained purely in simulation transfers well to the real world. Our planner enables the robot to rearrange a variable number of objects with a range of shapes and sizes using both push and pick-and-place skills.
\end{abstract}

\section{Introduction}
Robots operating in human environments must contend with many objects at a time. As such robotic multi-object manipulation and rearrangement has received much attention in the literature~\cite{cosgun-iros2011,chang2012interactive,gupta2012using, panda2013learning,dogar2014object,murali20206,paxton-corl2021-semantic-placement,liu-icra2022-structformer,zhu2020hierarchical}. The most recent of these works show excellent results in reasoning about novel objects from partial view sensory information~\cite{murali20206,paxton-corl2021-semantic-placement,liu-icra2022-structformer,zhu2020hierarchical,li2020towards, lin2022efficient, sharma2020relational}. However, robots using these approaches operate in a limited capacity manipulating individual objects one at a time.
In contrast, a pair of recent works have shown the ability for robots to reason about and manipulate multiple objects at once~\cite{wilson2020learning,suh2020surprising}. These works leverage image-based feedback controllers, that lack the level of semantic reasoning and explicit object grounding we desire for multi-step planning to logical goals. How can a robot plan to logical goals while still reasoning about a variable number of dynamically interacting objects?

Our work seeks to answer this question. We propose an approach that can explicitly predict inter-object relations in multi-object scenes, while also being able to plan to manipulate multiple objects at once.
We advocate for the use of logical relations for specifying goals as in~\cite{paxton-corl2021-semantic-placement,zhu2020hierarchical} as they provide a useful language for communication between robot and human. A human can easily construct a goal for tasking a robot by providing a conjunction of desired logical relations between objects in the scene. On the flip side, the robot can use its predictions of logical relations to communicate its belief over the current scene or future states it intends to achieve through manipulation. This contrasts with several recent approaches to rearrangement which provide images as goals to the robot~\cite{Qureshi-RSS-21}. Generating images for all desired goals requires a much higher burden on the user and in many cases would require the user to actively rearrange the scene, obviating the need for the robot!

To enable reasoning about manipulation effects simultaneously on multiple objects, we propose learning a latent state space in the form of a graph neural network (GNN). Our proposed relational dynamics GNN (\rdgnn{}) takes as input a segmented, partial view point cloud of the objects in the scene. It encodes this observation into the graph latent space, from which it can predict inter-object relations for both the current scene and future states given a sequence of actions. To enable these future predictions, we learn a dynamics function in the latent graph space. We then use our learned network to perform planning to achieve a desired relational goal.
Figure~1 illustrates how our learned model can be used for multi-step planning. Importantly, our planning framework can incorporate multiple distinct robot skills and produce multi-step task plans to achieve the specified goals.
Our planner performs diverse multi-object rearrangements including lifting and placing multiple objects at once, building and deconstructing towers, pushing objects to be in contact, and aligning objects spatially.

We leverage GNNs as their relational inductive bias enables better reasoning about multiple object interactions compared to alternative neural network structures~\cite{wilson2020learning}. Our experiments provide further evidence to this effect. We show that \rdgnn{} outperforms a similarly structured multi-layer perceptron operating directly on object pairs both in terms of planning success rate and predicting post-manipulation relations to enable successful planning. Crucially, using graph neural networks allows the robot to use the same model to reason about a variable number of objects. Further, by directly using partial view point cloud information as input, the robot can reason about objects of novel shape and size without access to explicit object models.

We further test the hypothesis that we can train \rdgnn{} using only the pre- and post-manipulation relational labels for supervision, in addition to the input point cloud and actions. Our extensive simulated and real world experiments show that using this relational supervision outperforms training to predict changes in object pose, coupled with an analytic approach to predicting relations from the object bounding boxes. Further, we show that training with both the relational and pose estimation losses provides no real benefit over training with relational losses alone.

\section{Related work}\label{sec:related-work}
Neural networks, including graph neural networks, have been applied to reason about spatial relationships and perform planning based on said reasoning~\cite{simeonov2020long,zhu2020hierarchical, paxton-corl2021-semantic-placement, liu-icra2022-structformer,yuan2022sornet}.
Paxton et al.~\cite{paxton-corl2021-semantic-placement} present a framework to reason about pairwise relations and plan to find an object placement that is physically feasible and satisfies the goal relations.
Liu et al.~\cite{liu-icra2022-structformer} present a transformer-based framework to manipulate objects into a configuration that satisfies the multi-object semantic goal relations encoded from natural language. This approach reasons about multi-object relations, but like~\cite{paxton-corl2021-semantic-placement} only moves a single object at a time via pick-and-place. It also requires added complexity incorporating language and loses the ability to operate explicitly on logical goals.
Yuan et al.~\cite{yuan2022sornet} present a framework to learn object feature embeddings incorporating single object semantics from RGB images for use in sequential manipulation tasks.
Simeonov et al.~\cite{simeonov2020long} leverage a GNN as a graph-attention network to select contact points and object transformations from point cloud observations for single-object manipulation.
Zhu et al.~\cite{zhu2020hierarchical} presents a grounded hierarchical planning framework for long-horizon planning manipulation tasks that leverages a symbolic scene graph to predict high-level plan actions and a geometric scene graph to predict low-level motions. Unlike our work, Zhu et al.~\cite{zhu2020hierarchical} do not examine multi-object dynamic interactions.
Lou et al.~\cite{lou2022learning} predict spatial relations between objects in clutter using GNNs to aid in finding better grasps, but do not model how relations will change post grasp. Furthermore, Driess et al.~\cite{driess2022learning} learn to predict multi-object interactions using graph nets, with supervised reconstruction for NERF-like embeddings. Unlike our proposed approach they do not predict object relations and learn and plan at a much finer time scale which makes their simulation-only experiments unlikely to transfer well to the real world. Biza et al.~\cite{biza2022factored} similarly examine learning object-oriented models of the world with pose estimation supervision. They show the ability to embed pose-based goals into a latent space, but do not explicitly reason about relations or manipulating multiple objects at once.
In~\cite{lin2022efficient} a GNN-based policy learns to perform multi-object rearrangement tasks including stacking and unstacking. However, the policy requires full object pose information and manipulates one object at a time.

Object stacking and unstacking tasks are challenging for robots to perform autonomously~\cite{bisk2018learning, paxton2019visual, zhang2019multi, jiang2012learning, jiang2013hallucinating} due to the difficulty in modeling the non-trivial contact dynamics and support relations of the objects being stacked.
Some recent works leverage GNNs for object stacking and unstacking tasks~\cite{li2020towards, lin2022efficient, sharma2020relational}.
Li et al.~\cite{li2020towards} leverage the GNNs to build relational reinforcement learning framework to help capture multi-object information in object stacking and unstacking.
However, this method does not consider generalization to objects of different shape and size, requires expert demonstrations, does not show real-world experiments, and uses only one primitive action.
Sharma and Kroemer~\cite{sharma2020relational} leverage GNNs to predict the feasibility of an action in object stacking and unstacking. They only consider preconditions, while we focus on leveraging multi-object dynamics to achieve logical goal relations. Furthermore, they require full 3D scene observations from multiple cameras, while we use the partial point cloud from a single camera.

Long-horizon planning has become an important problem for robot manipulation.
Task and motion planning (TAMP)~\cite{kim2019learning, garrett2017sample, kim2020learning, driess2020deep, garrett2020pddlstream, garrett2021integrated, liang-icra2022} defines a promising method to solve long horizon problems. TAMP approaches typically assume models of how objects and potentially their relations change. While learning has been used for various aspects of TAMP, no work has shown how to plan with multi-object dynamic interactions from point cloud data.
Simeonov et al.~\cite{simeonov2020long} propose an approach to object manipulations from point cloud data. They leverage a plan skeleton similar to us to solve long horizon planning problems. However, they do not reason about object relations and only manipulate one object with each action. Liang et al.~\cite{liang-icra2022} learn to plan with different skill primitives which sometimes include multi-object dynamic interactions. They perform multi-step skill planning using a heuristic graph search. However, they assume knowledge of object state and do not explicitly reason about object relations for learning.

\section{Planning to Goal Relations with GNNs}\label{sec:approach}
We assume our robot perceives the world as a point cloud $Z$ with \(N\) associated object segments $O_{i} \subset Z, i = 1,2,...,N$. The robot receives a goal,  $\gv = r_{1} \land r_{2}\land ...\land r_{M}; r_j \in \mathcal{R}$,  defined as \(M\) desired object relations.
$\gv$ represents the goal relation conjunction, $r_j$ represents each goal relation, and $\mathcal{R}$ represents the set of all possible relations.
Example relations in $\mathcal{R}$ include planar spatial relations such as ``object \textit{i} is in front of object \textit{j}'' or 3D relations such as ``object \textit{i} is above object \textit{j}'' and ``object \textit{i} is in contact with object \textit{j}.''
We assume the robot receives a plan skeleton $G = (\gv_{1}, \ldots, \gv_{H})$~\cite{lozano2014constraint, simeonov2020long} specifying the subgoals for each step in the multi-step plan of length $H$. We do not find this overly restrictive as several different approaches can generate appropriate plan skeletons~\cite{garrett2017sample, kim2020learning, driess2020deep,liang-icra2022}.

\begin{figure*}
    \centering
    \includegraphics[width=2\columnwidth]{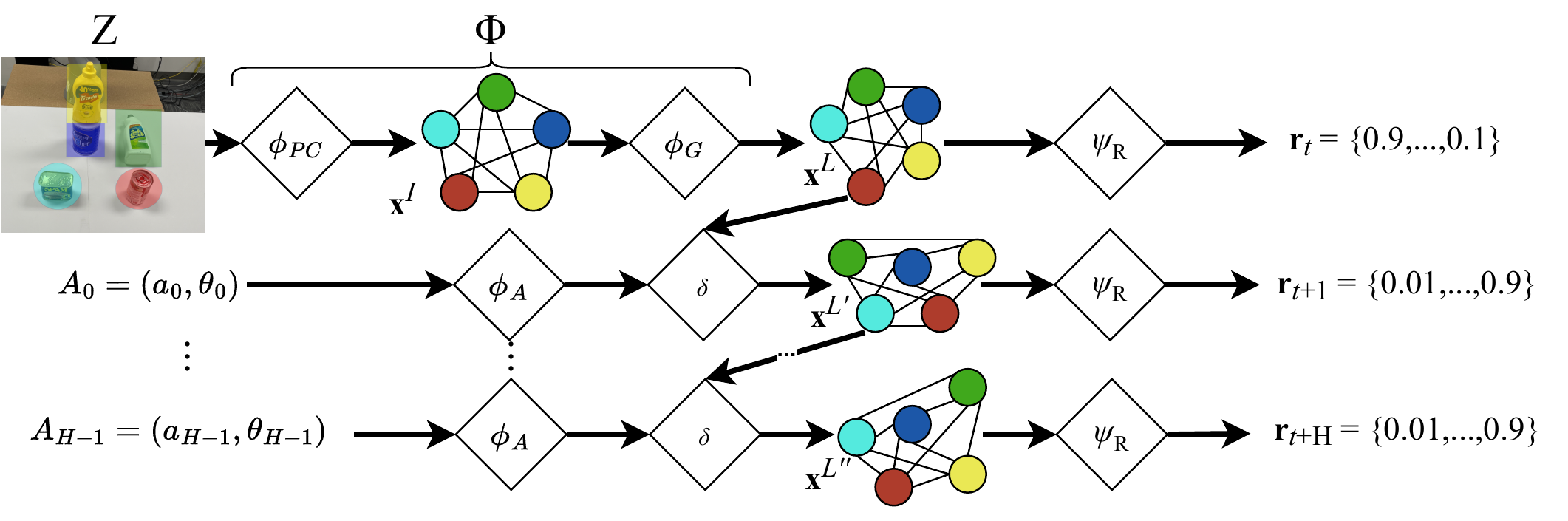}
    \caption{Overview of the components of \rdgnn{}. A point cloud encoder \(\pce\) transforms a segmented point cloud into \(N\) feature vectors acting as the node features for the fully connected input graph \(\ginput\). The graph encoder \(\graphenc\) transforms \(\ginput\) into a latent graph state embedding \(\glatent\). This latent graph in turn acts as input to a relational classifier \(\psi_R\) and dynamics function \(\delta\) along with actions encoded via \(\phi_{A}\).}
    \label{fig:graph_structure}\vspace{-18pt}
\end{figure*}

We provide our robot with a set of \(L\) parametric action primitives $\mathcal{A} = \{A_1, \ldots, A_L\}$ where \(A_l\) defines the discrete skill, which has associated skill parameters $\theta_{l}$. Example skills include a push skill~(\(A_l\)) with parameters~($\theta_{l}$) encoding the end effector pose and push length or a pick-and-place skill defined by the grasp and placement poses.

We define the robot's planning task as finding a sequence of skills and skill parameters $\tau = (A_{0}, \ldots, A_{H-1})$ that transforms the objects such that they sequentially satisfy each subgoal $\gv_{i}$ in the skeleton \(G\).
We propose learning a latent-space dynamics model~\cite{watter2015embed} for use in a planner to solve this task. The latent space model enables the robot to encode its partial view point cloud observations to a representation useful for planning.
Crucially we propose a novel graph neural network architecture to encode the latent space with a graph structure. This enables the robot to explicitly reason about a variable number of objects in the scene.

We learn an encoder to map observed segmented point clouds to the latent space \(\Phi: \mathcal{Z} \rightarrow \mathcal{X}\) and an associated decoder to predict inter-object relations from latent states \(\Psi: \mathcal{X} \rightarrow \mathcal{R}\).
To predict the forward state dynamics for planning we learn a forward dynamics function \(\delta: \mathcal{X} \times \mathcal{A} \rightarrow \mathcal{X}\).

We can now formally define our planning objective as maximizing the probability of achieving the goal relations with the following constrained optimization problem:
\begin{align}
  \argmax_{\tau=(A_{0}, \ldots, A_{H-1})} &\prod_{k=1}^{H} P(\rv_{k} = \gv_{k} | \xv_{k}) = \Psi(\xv_{k}) \label{eq:planning-obj} \\
  \texttt{subject to} \;\; &\xv_{k+1} = \delta(\xv_k, A_k) \; \forall k = 0, \ldots, H-1 \label{eq:dynamics-constraint}\\
               &\xv_{0} = \Phi(Z_0)  \label{eq:latent-grounding}\\
               &A_k \in \mathcal{A}  \;\;\;\;\;\;\;\;\;\;\;\;\;\;\;\,\forall k = 0, \ldots, H-1 \label{eq:skill-domain} \\ %
               &\theta_{min} \preceq \theta_k \preceq \theta_{max} \;\forall k = 0, \ldots, H-1 \label{eq:skill-constraints}
\end{align}
The constraints in this optimization problem encode the latent space dynamics (Eq.~(\ref{eq:dynamics-constraint})), grounding of the initial latent state from the observed point cloud Eq.~(\ref{eq:latent-grounding}), and constraints on the action parameters Eqs.~(\ref{eq:skill-domain}--\ref{eq:skill-constraints}).
We thus chain together predicted action effects decoding each state to predict the inter-object relations. Figure~1 visualizes planning with this model.

After solving this optimization problem the robot can execute the planned actions in the physical world. Our proposed network enables the robot to validate if it achieved its goal by computing \(\rv_k = \Psi(\Phi(Z_k))\). Where \(Z_k\) denotes the current point cloud observation.

We now provide a brief introduction to graph neural networks which are crucial building blocks for our proposed model. We then discuss the details of our specific relational dynamics graph neural network architecture visualized in Fig.~\ref{fig:graph_structure}. We follow this with a discussion of training our model before concluding this section with a description of our planning algorithm.

\subsection{Graph Neural Networks}
We define a directed graph $G = \{V, E\}$ with nodes $V = \{\vv_i\}$ and edges as \(E = \{\ev_{\ij}\}\) where each \(\vv_i\) and \(\ev_{\ij}\) is a feature vector for node \(i\) or the edge from \(i\) to \(j\) respectively. We seek to encode information associated with this graph into a neural network; following~\cite{graph_nets} we can reason about our graph network operations in terms  of message passing in the graph, where a single graph net layer of \emph{update} and \emph{aggregation} functions performs one round of message passing between neighbors in the graph. By constructing multiple graph layers, information from nodes across the graph can propagate in the form of deeper and deeper features. %

Update functions transform individual node or edge features. We use feed-forward multi-layer perceptrons as update functions in this paper. We denote node updates as $\vv_{i}^{\prime} = f_{n}(\vv_{i})$ and edge updates as $\ev_{\ij}^{\prime} = f_{e}(\ev_{\ij})$. Aggregations take inputs from multiple parts of the graph and reduce them to a fixed feature length, thus enabling consistent output feature dimensions from a variable input size. We denote a message from node \(i\) to node \(j\) as \(\mv_{\ij} = (\vv_i \oplus \vv_j \oplus \ev_{\ij})\) and define our message update functions as \(\mv_{\ij}^\prime = f_{m}(\mv_{\ij})\). Here \(\oplus\) denotes vector concatenation. To define our aggregation functions we introduce an intermediate variable \(\yv_i = \frac{1}{|\mathcal{N}(i)|}\sum_{j \in \mathcal{N}(i)}(\mv_{ji}^{\prime})\) which takes the average of all messages incoming to node \(\vv_i\) denoted as those coming from nodes in node \(i\)'s neighborhood \(\mathcal{N}(i)\).
Using this we can define our node aggregation function as $\vv_{i}^{\prime} = g_v(\vv_{i} \oplus \yv_i)$ and the edge  aggregations as \(\ev_{\ij}^{\prime} = g_{e}(\vv_i \oplus \vv_j \oplus \yv_{i} \oplus \yv_{j} \oplus \ev_{\ij})\)
where \(g_v(\cdot)\) and \(g_{e}(\cdot)\) define MLPs.
This edge aggregation thus concatenates and then transforms the features associated with the two neighboring nodes and the messages passing between them.
For more details on graph nets including alternative aggregation functions see~\cite{graph_nets}.

\subsection{Learning Relational Dynamics with GNNs}
We now turn our attention to our relational dynamics graph neural network, \rdgnn{}, which takes as input the segmented object point cloud and a candidate action and predicts the current and post-manipulation inter-object relations.
We discuss the network in terms of several different components: the encoder, the latent graph dynamics function, and the relational output classifier.
An overview of these different components and how they connect to one another is visualized in Figure~\ref{fig:graph_structure}.

Our encoder \(\Phi\) can be decomposed into two sub-networks: the point cloud encoder and the latent graph encoder.
Our point cloud encoder \(\pce(Z)\) operates on each of the \(N\) point cloud segments, \(O_i \subset Z\), converting the variable size input point cloud to a fixed-length feature vector \(\vv_i\). This feature vector will act as the node feature to our input graph to the GNN. We use \emph{PointConv}~\cite{wu2019pointconv} as the backbone of our point cloud encoder to output a feature of length 128.

Given the output of our point cloud encoder, we define our input graph $\ginput = (V^{I}, E^{I})$ with nodes \(V^{I} = \{\pce(O_i) \oplus k \}\) where \(k\) denotes a one-hot encoding providing a unique identity label for each node.
To improve the generalization ability we randomly generate the object IDs during training~\cite{cui2022positional} over a range larger than the highest number of objects expected to be seen at deployment.
We chose 16 in the paper since it was all we needed for our experiments, although it could be set higher for other applications.

We define edges to and from all node pairs in the graph creating a fully-connected, directed input graph. We set all input edge feature \(\ev_{\ij} \in E^{I}\) to be empty. This topology enables message passing between all nodes, but provides no explicit edge features as input for learning.

We use our graph encoder to transform our input graph, \(\ginput\) to a latent graph $\glatent = \graphenc(\ginput)$. Here \(\graphenc(\cdot)\) represents a layer of graph message passing and aggregation as defined in the previous section.
We use our latent graph embedding as input to two sub-networks: our relational classifier, \(\rv = \psi_R(\glatent)\) and our latent graph dynamics function \(\glatentprime = \delta(\glatent, A)\). %

We construct our relational classifier as an MLP that operates on a pair of nodes and their associated edges from \(\glatent\), taking the form of an edge aggregation network \(\rv_{\ij} =  \psi_{R}(\vv_i, \vv_j, \yv_{i}, \yv_{j}, \ev_{\ij})\).
We predict relations for all object pairs by running this classifier for each pair of nodes in the graph as a form of graph convolution. While some relations may be mutually exclusive, in general the spatial relations are independent of one another, necessitating individual binary classifiers and not a softmax-based multi-class classifier. Note we never specify mutually exclusive goal relations.

We additionally examine learning to predict the object pose (defined as its centroid and bounding box orientation in simulation) for all objects in the scene. To this end we learn a pose regressor \(\xv_i = \psi_{P}(\glatent)\) which we train using a node aggregation network with an output MLP with 3 outputs encoding position and 6 encoding orientation as in~\cite{zhou-cvpr2019continuity}.

The final piece to define is our latent graph dynamics function \(\glatentprime = \delta(\glatent, A)\).
Recall that \(A\) defines the action (skill) including its skill parameters being evaluated through the dynamics. We encode any discrete skill variables (e.g. object identity) using a one-hot-encoding for use as input into the network.
We pass this action through an action encoder \(A^{\prime} = \actionenc(A)\) which we implement as an MLP.
We build separate node \(\delta_{v}(\cdot)\) and edge \(\delta_{e}(\cdot)\) dynamics functions which respectively take as input the node or edge features of the latent graph concatenated with the encoded action. As output they predict the change in graph features \(\Delta \vv_i^{L}, \Delta \ev_{\ij}^{L}\). Given these definitions we define our graph dynamics functions as
\(\vv_i^{L^\prime} = \vv_i^{L} + \delta_{v}(\vv_i^{L} \oplus \actionenc(A))\)
and
\(\ev_{\ij}^{L^\prime} = \ev_{\ij}^{L} + \delta_{e}(\ev_{\ij}^{L} \oplus \actionenc(A))\).
We incorporate multiple skills by learning a separate dynamics functions for each skill, using the same shared latent space.

\begin{figure*}[ht]
    \centering
    \includegraphics[width=1.98\columnwidth]{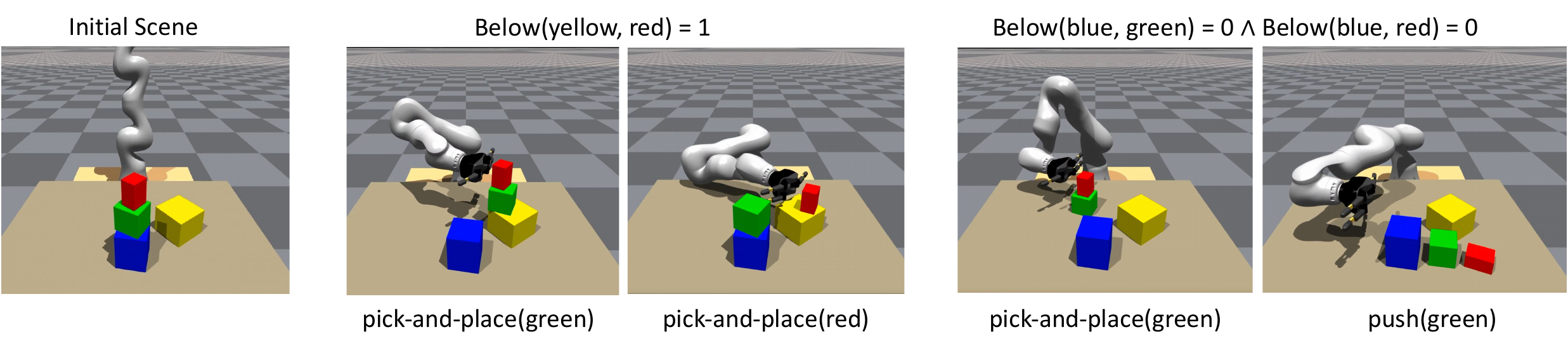}
    \caption{For the same initial scene (left) we show different valid states found by our planner and model for two different goal settings.
    For the first goal relation, the robot can either pick the green object or the red object to place atop the yellow object.
    For the second goal relation, the robot can either push the green object or pick-and-place the green object to deconstruct the towers.
    Here we label the objects using color instead of object id to make the figure easier to interpret.
      }
    \vspace{-18pt}
    \label{fig:plan_distribution}
\end{figure*}

\subsection{Multi-step Loss Functions and Model Training}
We train our model end-to-end using a combination of loss function terms. The first term defines the classification (cross-entropy) loss of predicting the current relations from the currently observed point cloud \(\mathcal{L}_{\text{REL}} = \sum_{t=1}^{H}\texttt{CE}(\rv_t, \hat{\rv}_t = \psi_R(\Phi(Z_t)))\).
Our second loss component provides regularization that the predicted, post-manipulation latent graph should match the latent graph encoded from the point cloud observation after executing the associated action~\cite{watter2015embed} \(\mathcal{L}_{\text{DYN}} = \sum_{t=1}^{H}\sum_{i=0}^{H-t-1}||\glatent_{t+i} - \glatentprime_{t+i}||_2^2\). Where \(\glatent_{t+i} = \Phi(Z_{t+i})\) defines the latent graph encoded directly from an observed point cloud and \(\glatentprime_{t+i+1} = \delta(\Phi(Z_{t+i}), A_{t+i})\) defines the predicted latent graph state through the learned dynamics function.
We recursively use our $\delta$ functions to predict the latent state for all time steps in the executed plan.

Our final loss term operates on the post-manipulation relations predicted via the latent graph dynamics, \(\mathcal{L}_{\text{REL}^{\prime}} = \sum_{t=1}^{H}\sum_{i=0}^{H-t-1}\texttt{CE}(\rv_{t+i}, \hat{\rv}^{\prime}_{t+i})\). %
We make the distinction between \(\hat{\rv}_{t+i}\) and \(\hat{\rv}_{t+i}^{\prime}\) explicit, where \(\hat{\rv}_{t+i} = \psi_R(\glatent_{t+i})\) and \(\hat{\rv}_{t+i}^{\prime} = \psi_R(\glatentprime_{t+i}) \). This operates in an analogous multi-step fashion to the latent dynamics regularization.

For pose estimation we define an L2 loss on the current and predicted object poses in an analogous manner replacing \(\psi_R\) with \(\psi_P\).
We examine the effect of this loss in our experiments.

\subsection{Planning Algorithm Implementation}
Any number of methods can solve the optimization problem defined in Eq.~(\ref{eq:planning-obj}). For this paper we use the cross entropy method (CEM)~\cite{rubinstein2004cross,kobilarov2012cross}. CEM is a derivative-free, sampling-based procedure that begins with an initial Gaussian distribution over the decision variables $\mathcal{N}(\tau \mid \mu_0, \Sigma_0)$. We generate a set of samples $S = \{\tau_i \sim \mathcal{N}(\tau \mid \mu_0, \Sigma_0)\}$, evaluate each one under the specified cost function, and select the top-$k$ low-cost samples $S_{\texttt{topk}}$. We recompute the mean and covariance for the current iteration as $\mu_1 = \texttt{mean}(S_{\texttt{topk}})$ and $\Sigma_1 = \texttt{cov}(S_\texttt{topk})$, and then proceed to the next iteration. We repeat this process for $K$ iterations and select the final mean $\mu_M$ as the result. Our optimization problem requires a mixed discrete-continuous optimization due to having to find a sequence of skills (discrete) and the parameters to those skills that are themselves mixed discrete-continuous.
Because our robot has a small number of skills, we independently search for each discrete skill and select the one with the lowest cost (highest success probability).
To improve the numerical stability we minimize the log of Eq.~(\ref{eq:planning-obj}) instead of Eq.~(\ref{eq:planning-obj}).
Given the subgoal skeleton we can greedily solve the continuous and discrete action search for each step of the plan. We propagate the predicted latent state, \(\xv_{t+1}\) resulting from running action \(A_t\) as initial state for the next plan step.

\section{Experiments \& Results}\label{sec:experiments}
We now describe the training data collection and experimental validation for our approach to learning and planning with \rdgnn{}.
In our experiments we examine the following relations: \texttt{left}, \texttt{right}, \texttt{behind}, \texttt{in-front}, \texttt{above}, \texttt{below} and \texttt{in-contact}.
We get the \texttt{in-contact} relation directly from simulation and we define other relations following Paxton et al.~\cite{paxton-corl2021-semantic-placement}.
We train and evaluate our model on multi-object rearrangement tasks using pushing and pick-and-place skills.
We conduct experiments in simulation and on a physical robot manipulating both blocks and YCB objects~\cite{calli2015ycb}.

\noindent \textbf{Dataset collection in simulation:}
We conduct large scale data collection using the Isaac Gym simulator~\cite{isaacgym}.
We collect a dataset by generating scenes with a variable number of cuboid objects of random size with arbitrary pose.
Scenes contain objects in either one or two vertical stacks.
We then execute a random push or pick-and-place action on one of the objects in the scene. We record the partial view point cloud before and after the manipulation, the executed action, and the ground truth relations between all object pairs in the scene. We collected a total of 39,600 push and pick-and-place attempts.
Fig.~\ref{fig:plan_distribution} shows an example scene with various pushing and pick-and-place actions and outcomes from the simulator.

\noindent \textbf{Baseline Approaches:} We implement several baselines for comparison to our proposed model \emph{\rdgnn{}}.
\emph{PointConv Relations (PCR)}: defines a \emph{PointConv} based network that takes in a pair of segmented objects and predicts their relations without any GNN, similar to the relational classifiers in~\cite{paxton-corl2021-semantic-placement}.
\emph{Pairwise MLP Relational Dynamics (MLP)}: predicts relations and dynamics for pairs of objects using an MLP instead of a GNN to construct the latent space and dynamics.
\emph{Direct Pose Dynamics GNN (DPD-GNN)}: uses a GNN to predict the pose for each object conditioned on a chosen action. We use an analytic relational classifier to predict relations from the predicted poses and their associated bounding boxes.
\emph{Pose Estimation GNN (PE-GNN)}: We replace the relational output heads on our model with pose estimation regressors. We again use analytic relational classifiers for evaluation.
\emph{Combined Relational Dynamics and Pose Estimation (RD-PE-GNN)}: This combines our model with the pose estimation regressor for both the current and next time step.
\emph{Relational Dynamics without Latent Regularization (RD-GNN-w/o-LR)}: We train a version of our model without using the \(\mathcal{L}_{\text{DYN}}\) loss.

\begin{figure*}
    \centering
    \includegraphics[width=0.66\columnwidth,clip,trim=3mm 0mm 10mm 6mm]{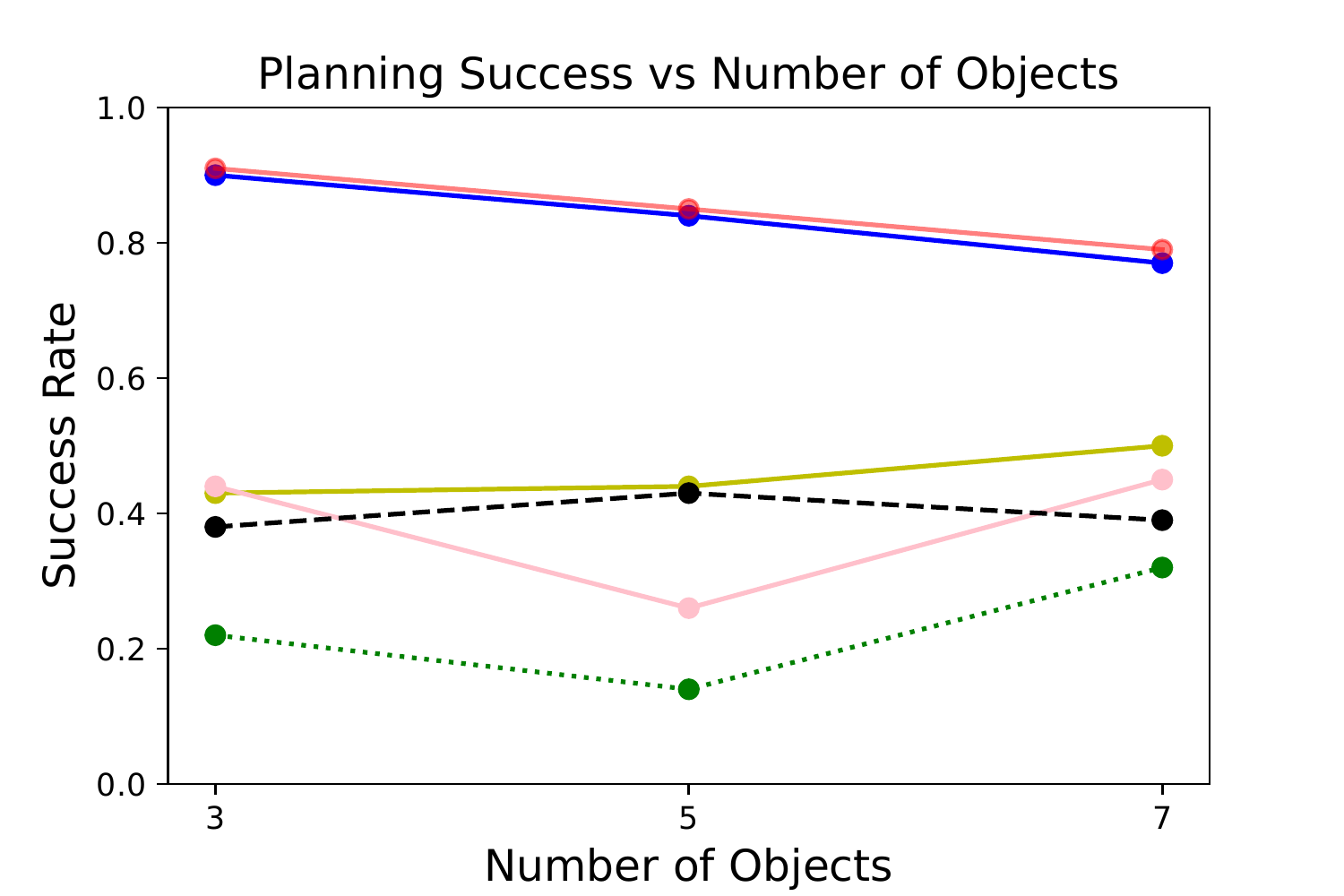} \label{fig:comparison_obj_num}  %
    \includegraphics[width=0.66\columnwidth,clip,trim=3mm 0mm 10mm 6mm]{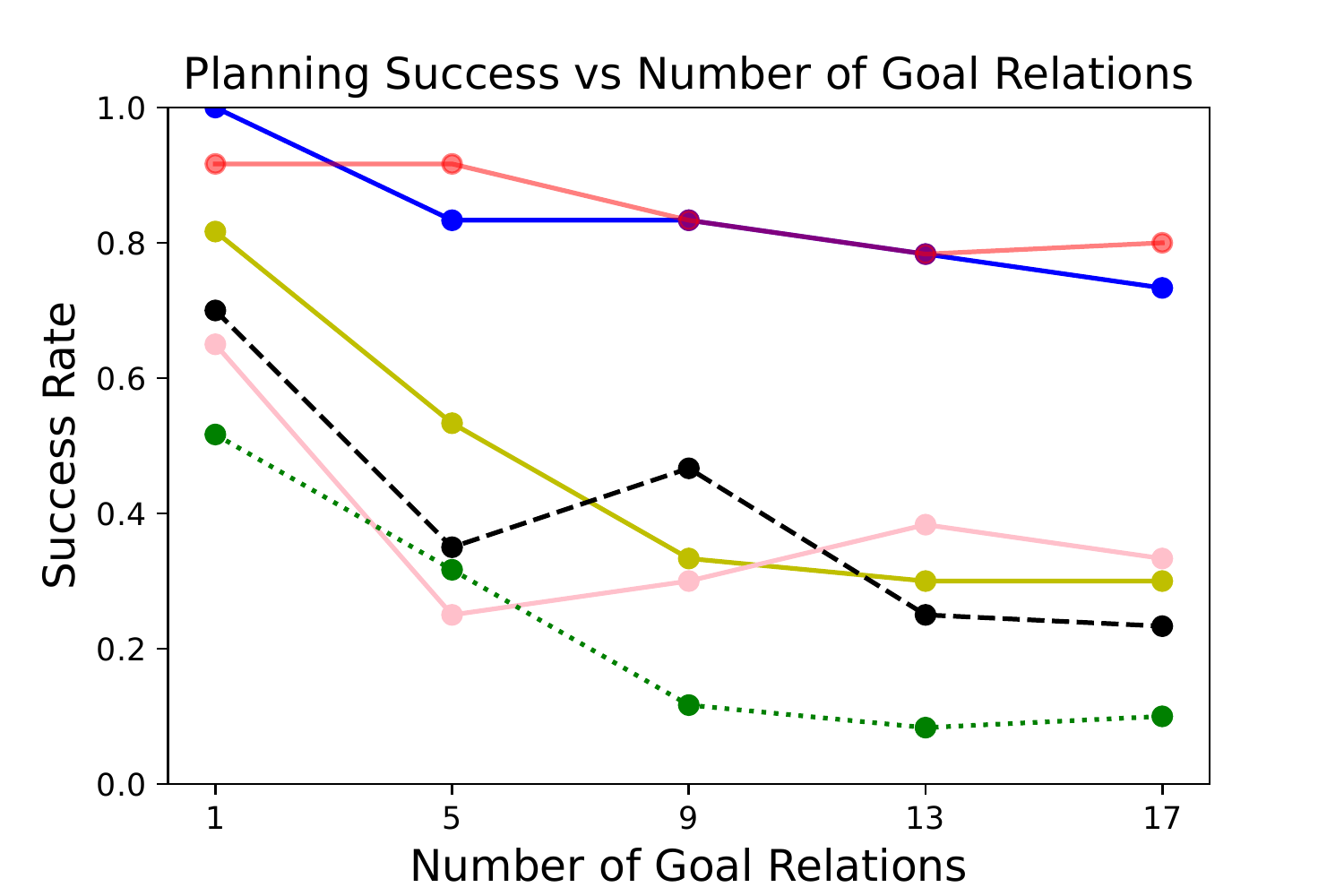} %
    \includegraphics[width=0.66\columnwidth,clip,trim=3mm 0mm 10mm 6mm]{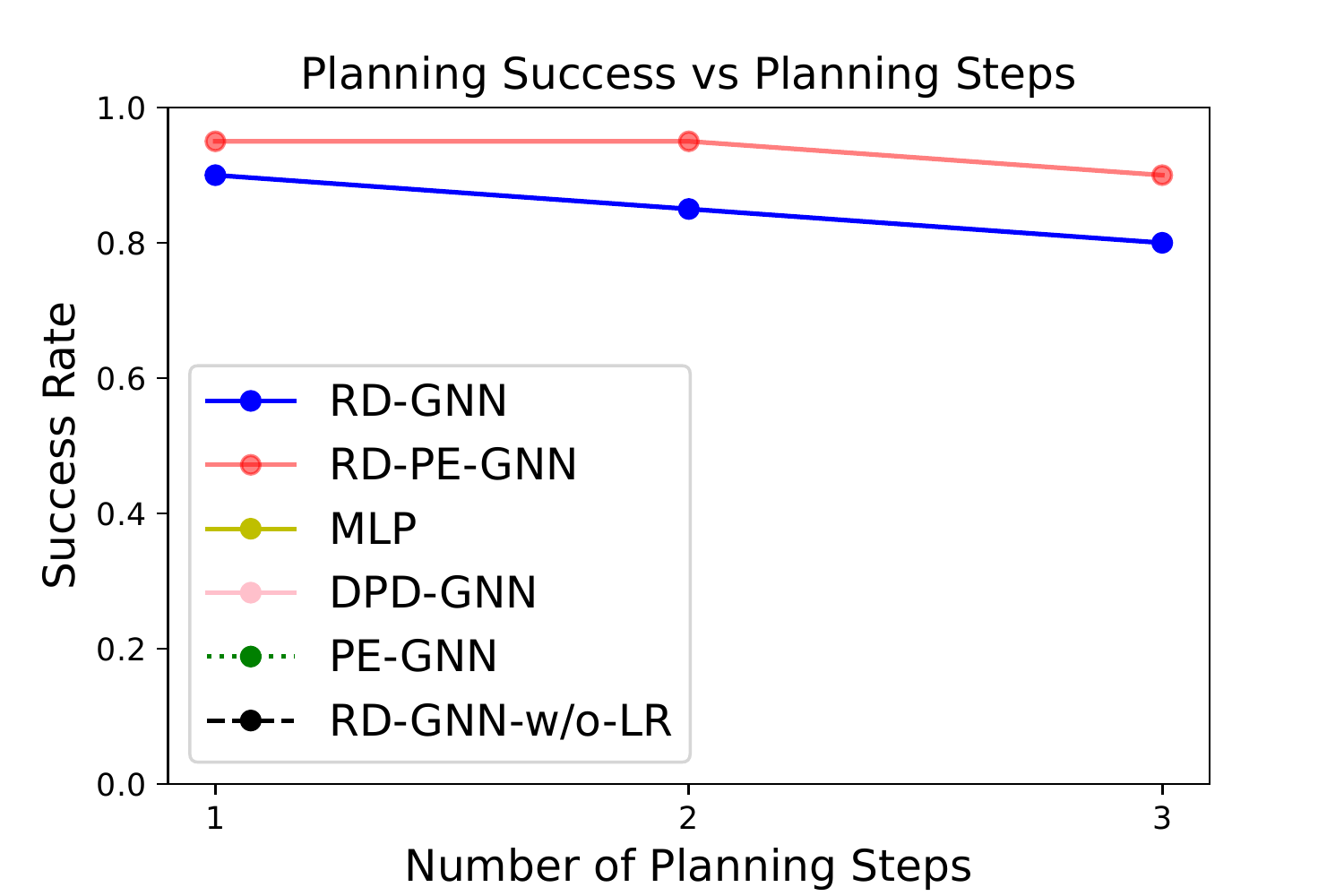}
    \caption{Comparing planning success rate of the different models as a function of (left) the number of objects in the scene, (middle) the number of relations specified in the goal, and (right) the number of steps. The legend applies to all three plots. We see that RD-GNN and RD-PE-GNN achieve comparable performance while significantly outperforming the baseline models. The success rate drops for all models as we specify more relations in the goal. Even when fully constrained the top performing models achieve high success rates.}\vspace{-18pt}
    \label{fig:planning_success_comparison}
\end{figure*}

\noindent\textbf{Predicting and Detecting Inter-Object Relations:} We first examine the efficacy of our model in correctly predicting which relations will be present after executing a specified action. Then we examine the ability of our model to detect inter-object relations for objects in the observed scene post manipulation.
We show here the prediction accuracy between the predicted relations and the ground truth relations post-manipulation.
On our simulation test data across 300 skill executions, the average prediction F1 score across all relational classifiers predictions for \rdgnn{} is 0.906, MLP is 0.678, RD-PE-GNN is 0.879, DPD-GNN is 0.319, PE-GNN is 0.133, and RD-GNN-w/o-LR is 0.693.
We find that the \rdgnn{} and RD-PE-GNN perform comparably and outperform other baselines in terms of prediction F1 score.
We next show the F1 score between the detected and ground truth relations post manipulation.
For our simulation test data, the average F1 score of the post manipulation relational classifiers for \rdgnn{} is 0.974, MLP is 0.971, RD-PE-GNN is 0.977, PointConv is 0.985, PE-GNN is 0.899, and RD-GNN-w/o-LR is 0.977.
We find that our approach performs comparable to the PCR baselines, MLP, RD-PE-GNN, and RD-GNN-w/o-LR. 
Furthermore, our approach outperforms PE-GNN in terms of detection F1 score.
Note the PCR method alone cannot be used for planning dynamic interactions, which is the main focus of this work. %

\noindent \textbf{Planning to Desired Goal Relations:} We now examine the ability of our model to plan to desired goal relations with a single action step.
We ran 20 planning trials containing varying numbers of objects and goal relations using each model in simulation. We only used pushing tasks for large-scale statistics. 
Fig.~\ref{fig:planning_success_comparison} shows that our model, RD-GNN, as well as our model with pose estimation, RD-PE-GNN, dominate all competitors.
Fig.~\ref{fig:plan_distribution} shows a variety of successfully executed single-step plans using RD-GNN. Notably we generate diverse plans for the same goal and initial setting.

\begin{figure}[ht]
  \centering
  \vspace{-5pt}
  \includegraphics[width=0.98\columnwidth,clip,trim=15mm 2mm 10mm 8mm]{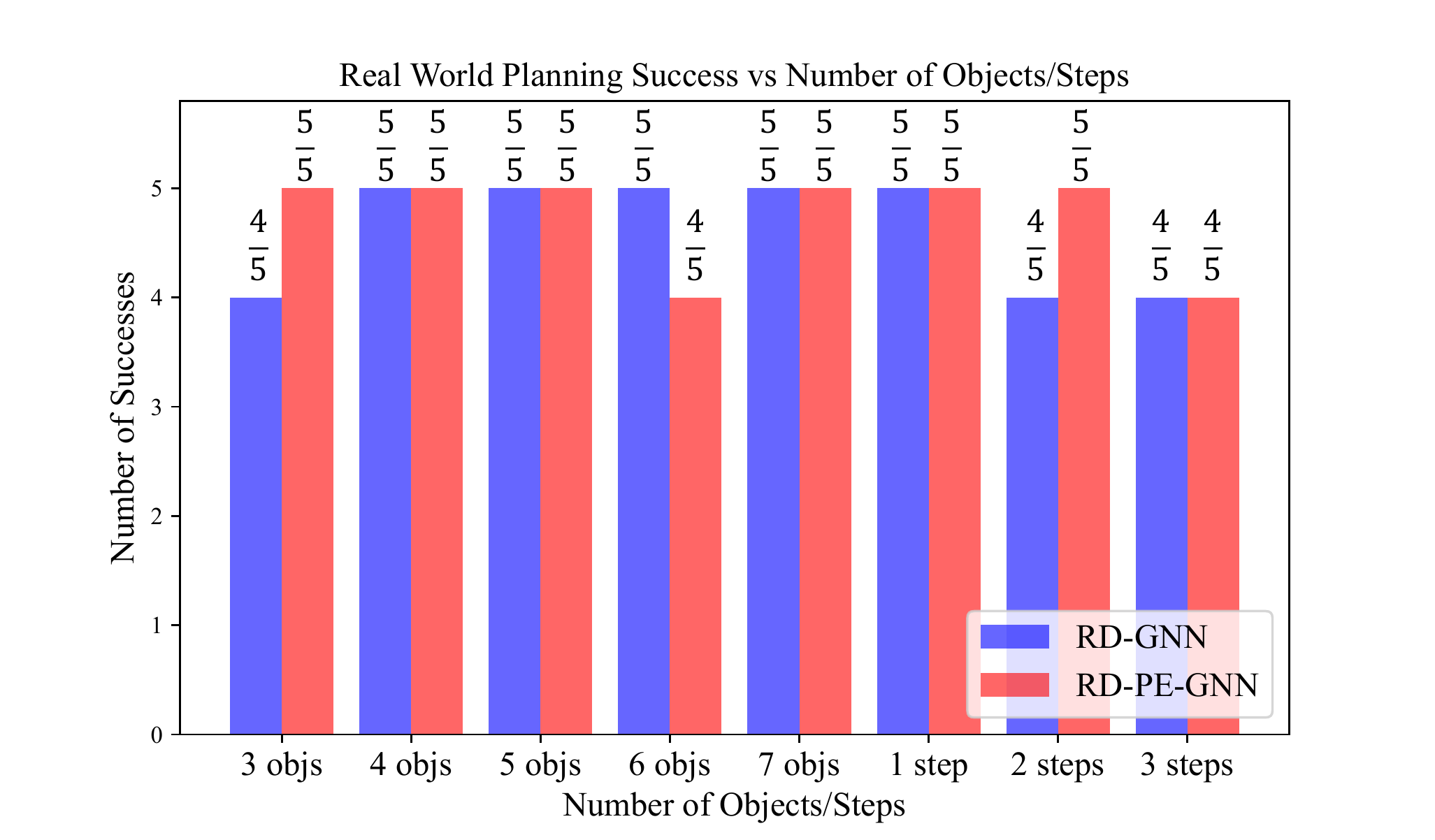}  %
  \caption{Number of successes on real-world YCB object manipulation tasks. We compare results for a varying number of objects and varying plan horizon length as denoted by the horizontal labels.}\vspace{-5pt} \label{fig:real_all}
\end{figure}

We now turn our attention to multi-step planning focusing on only RD-GNN and RD-PE-GNN as the best single step performers. We show planning success rates for plans ranging in length from 1 to 3 planning steps in simulation in Fig.~\ref{fig:planning_success_comparison} (right).
We use four objects for the multi-step test in simulation and the real world.
We see that for both the \rdgnn{} and RD-PE-GNN approaches, the success rate drops with plan length. Both models achieve high success rates.

We show planning success rate for real-world experiments in Fig.~\ref{fig:real_all}.
We ran test trials using YCB objects with 5 trials for each setting of varying number of scene objects or plan steps.
For all the real-world experiments, we use 5 relations in the goal.
Our results verify that our method transfers to real-world settings without any fine tuning and generalizes to real-world objects when trained only on cuboids in simulation. Fig~1 illustrates real-world multi-step plan execution.

\section{Conclusion}\label{sec:discussion}
We presented an approach to multi-object manipulation based on learning relational classifiers via graph neural networks.
We showed through extensive simulated and physical robot experiments that the relational inductive bias of the graph net provided improved planning success.
We can attribute this to better accuracy when predicting both inter-object relations and latent space dynamics.
Further we showed inter-object relations provide a better source of supervision for training our model for planning than using a pose estimation loss.

While we find our results quite exciting, several limitations exist in the approach as currently formulated.
On a theoretical level we have no proof that the relations we use in training our representation provide a sufficient basis for predicting all inter-object interactions of interest. Currently we only have empirical results to show they seem to work well. As an additional training issue, we only trained with block-shaped objects, while this proved sufficient for demonstrating the benefits of relational graph dynamics, we do not capture detailed shape information that robots must reason about for more complicated tasks and interactions.
At a low-level we have not closely integrated the motion planner we use for skill generation with the higher-level CEM planning. This causes the robot to sometimes reject samples that fail to generate motion plans, decreasing efficiency and coverage of our planner.
For the multi-step planning, we do not use replanning which requires high accuracy of the latent dynamics model. We think this mechanism will not generalize to very long horizons like 20 steps.
So we plan to do replanning for future works.

Overall, our approach provides the first example of predicting multi-object manipulation sequences using learned relational dynamics from partial view point clouds. We leverage these predictions for planning and executing multi-object dynamic rearrangements on a physical robot.

\section*{Acknowledgments}
The authors thank Mohit Sharma, Chris Paxton, and Mohanraj Devendran Shanthi for useful discussion.
This work was partially supported by NSF Award \#2024778, by DARPA under grant \mbox{N66001-19-2-4035}, and by a Sloan Research Fellowship. 
\bibliographystyle{IEEEtran}
\bibliography{references}

\clearpage
\newpage
\pagenumbering{arabic}%
\renewcommand*{\thepage}{A\arabic{page}} %
\appendix
\section{Appendix}
\subsection{Spatial Relations Definitions}
We define 7 relations (\texttt{left}, \texttt{right}, \texttt{behind}, \texttt{in-front}, \texttt{above}, \texttt{below} and \texttt{in-contact}) for our experiments.

\textit{in-contact}(A,B) = 1 if A and B is in contact, we can get this from the simulation directly.
We get other relations based on Paxton et al.~\cite{paxton-corl2021-semantic-placement}.
And since all the labels are based on Isaacgym simulation, so we assume no penaltry and other cases that does not satisfy simulation kinematics and dynamics.
Since sometimes the object will fall over from the stack and then might be off the camera view.
So in this case, we will detect the relations between all other objects and then manually set this object as the off camera view relations.
For visualization purpose, we redefine the relations in the camera's perspective.

\subsection{Neural Network Details}
We use Sigmoid activation function for the classification output layer. Otherwise, we use ReLU for all activation functions.
For the graph encoder, We pass the one-hot vector through one fully connected layer mapping it to a 128 dimensional vector.
We use node and edge MLPs with one hidden layer each of width 64 outputting latent graph node and edge features of 128 dimensions.
For the relational classifier in our experiments, \(\psi_{R}(\cdot)\) has one hidden layer of width 64 in the aggregation output before a final output layer of width equal to the number of relations (7 in this work)  with a sigmoid activation to create a binary classifier for each relation. For the pose estimation head, it has one hidden layer of width 64. For the action encoder, it has one hidden layer with width 128 and that outputs a 128 dimensional action encoding.

\subsection{Implementation Details}

We implemented the push skill based on the push direction and push distance.
Our action primitive encodes which object to push, which direction it will push, and what distance it will push in this direction. The initial end-effector pose is computed as a fixed offset from the object point cloud centroid along the negative push direction.

We implemented the pick-place skill based on the object point cloud's current centroid and major axis (as an approximate pose) and the desired placement pose.
The action primitive encodes which object to pick and the planar displacements between the pick and  place poses. We use a bounding-box heuristic for pick pose candidates, where we choose from poses that align the end-effector to the axes of a bounding-box for the selected object geometry. 
To highlight the versatility of our approach, we created an additional dataset using a different grasping skill that constrains the change in end effector orientation.
For both skills we use a one-hot encoding to denote the object to be manipulated.

\subsection{Point Cloud Segmentation}
In simulation we use the ground truth segmentation masks provided by the IsaacGym simulation.

For real-world segmentation we use a joint color and depth based segmenter based on superpixel algorithms~\cite{felzenszwalb2004efficient}. We extended the method to also use depth information in its distance computation.

\begin{figure}[ht]
  \centering
  \includegraphics[width=0.98\columnwidth,clip,trim=1mm 2mm 7mm 1mm]{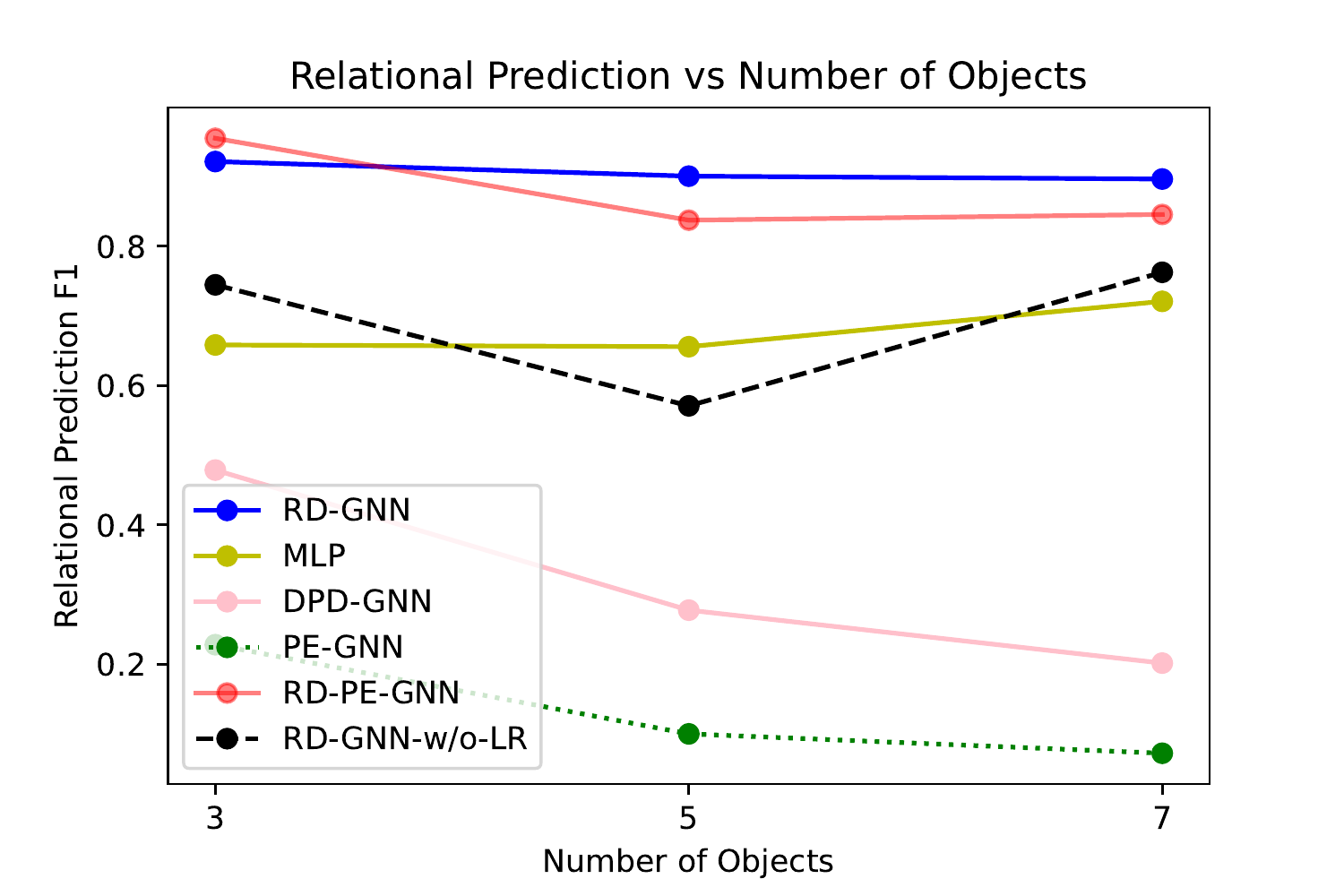}  %
  \caption{Comparing prediction F1 score of the different models as a function of the number of objects in the scene. We see that RD-GNN and RD-PE-GNN achieve comparable performance while outperforming the baseline models for the prediction F1 score.} \label{fig:relational_F1}
\end{figure}

\begin{figure}[ht]
  \centering
  \includegraphics[width=0.98\columnwidth,clip,trim=1mm 2mm 7mm 1mm]{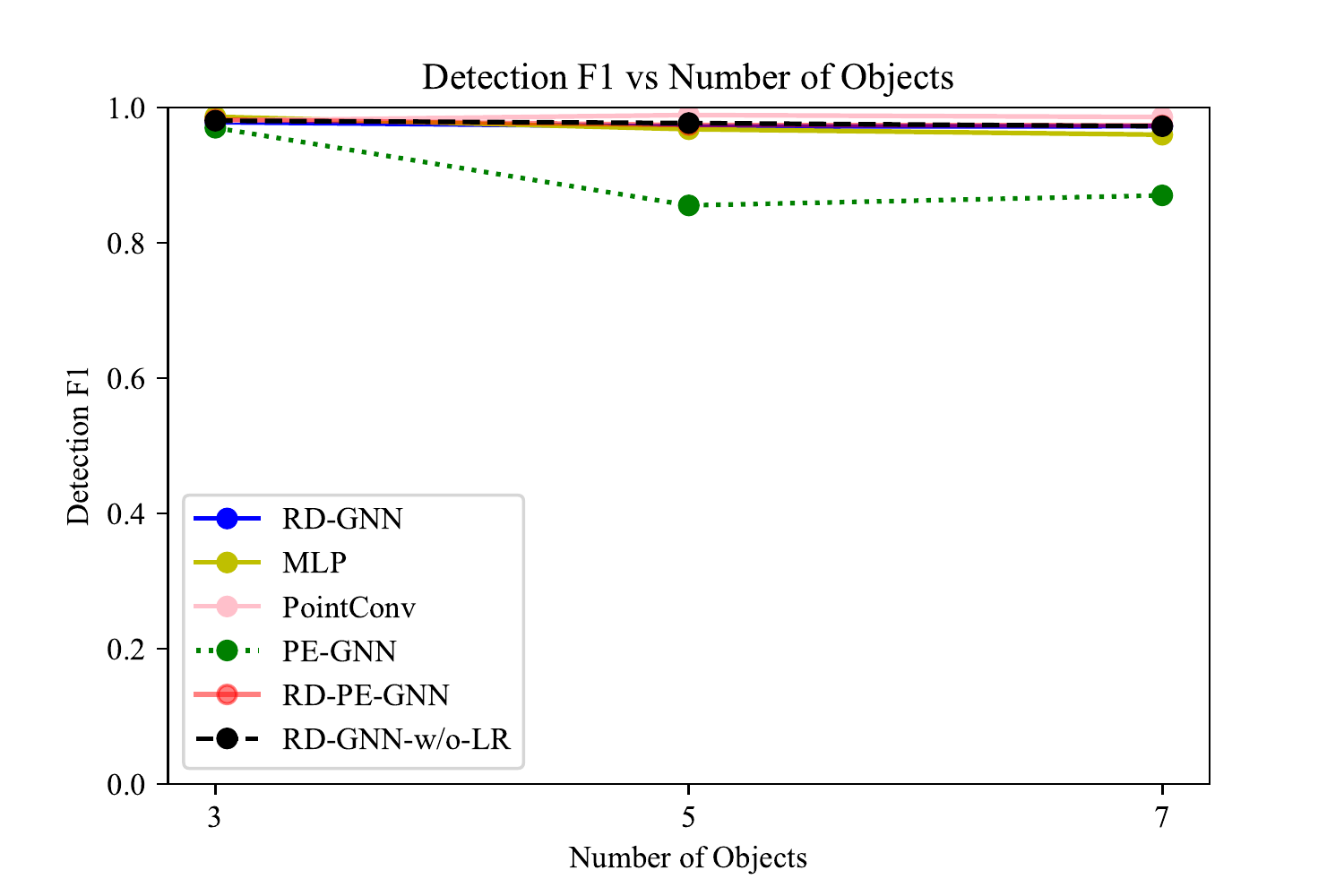}  %
  \caption{Comparing detection F1 score of the different models as a function of the number of objects in the scene. we can find that our approaches perform comparable to the PCR baselines, MLP, RD-PE-GNN, and RD-GNN-w/o-LR baselines while PE-GNN performs poorly in the detection F1 score.} \label{fig:Detection_F1}
\end{figure}

\subsection{CEM Planning}
For the details of the continuous part of the planning algorithms. We focus on the search of the x and y directions. For the z direction, we keep it the same height as the initial pose of the object. We choose $\mu_0$ as (0, 0). Since the robot has different reachable space for different skills, we chose $\Sigma_0$ as (0.05, 0.3) for the push skill and $\Sigma_0$ as (0.3, 1.1) for the pickplace skill.
To make sure the action is in the reachable workspace of the robot, we also need to adapt this sampling range $\Sigma_0$ based on the initial pose of the object.
For each iterations, we sample 200 actions and chose the top 3 samples. We choose total iterations $M = 2$ in our implementation. 

When we execute an action from planning with CEM, instead of executing the $\mu_M$ as commonly done \cite{rubinstein2004cross}, we randomly select an action within three standard deviations of the mean. We found this improves the low-level motion planner success rate, since the mean action may be difficult to reach given the robot's limited reachable workspace.

\subsection{Model Training Details}
All our models are trained on a standard workstation.
We set our training batch size = 1 due to the limited GPU resources. We train the model using the ADAM optimizer with an initial learning rate of 0.0001.
We use data from 39,600 trials with both push and pickplace skills to train our model.

\subsection{Extra Simulation Results}
We show extra results detailing  F1 scores for relational prediction and detection in Fig.~\ref{fig:relational_F1} and Fig.~\ref{fig:Detection_F1}.

\end{document}